\documentclass[11pt]{article}

\usepackage[final]{acl}

\usepackage{booktabs}
\usepackage{multirow}

\usepackage{CJKutf8}

\usepackage{xcolor}
\usepackage{colortbl}

\usepackage{times}
\usepackage{latexsym}

\usepackage[T1]{fontenc}

\usepackage[utf8]{inputenc}

\usepackage{microtype}

\usepackage{inconsolata}

\usepackage{graphicx}

\title{Isolating the Language Shell: A Content-Controlled Study of Multilingual Reasoning}

\author{
 \textbf{Hoejoon Kwon\textsuperscript{1}},
 \textbf{Jinhee Jang\textsuperscript{1}},
 \textbf{Byeonggeuk Lim\textsuperscript{2}}\thanks{Currently at KETI, banggeuk@keti.re.kr},
 \textbf{YoungBin Kim\textsuperscript{2}}
\\
\\
 \textsuperscript{1}Department of Artificial Intelligence, Chung-Ang University, \\
 \textsuperscript{2}Graduate School of Advanced Imaging Science, Multimedia \& Film, Chung-Ang University
\\
  \texttt{\{chriskwon9, jinheejang, bangguek, ybkim85\}@cau.ac.kr}
}

\begin{document}
\maketitle
\begin{abstract}
Multilingual reasoning distillation requires choosing the language in which
teacher reasoning is presented, yet existing recipes often entangle
linguistic form with differences in generation, translation, and filtering.
We construct content-controlled supervision by rendering the same verified
English trajectories into English, translated target-language, and
language-mixed shells, and fine-tune two pretrained students in three target
languages, evaluating them on multilingual mathematical reasoning at two
difficulty levels. Even with the
underlying reasoning content held fixed, we find that shell choice can
change downstream accuracy, with effects varying across students, languages,
and difficulty levels. Pre-training compatibility does not consistently track these outcomes, and even when our tests find no clear accuracy difference between shells,
their output-language behavior can still differ substantially. These results suggest that the supervision shell is a distinct design choice in multilingual reasoning
distillation.
\end{abstract}

\section{Introduction}

Reasoning has become a central capability of modern Large Language Models
(LLMs), with reasoning-oriented training and inference-time scaling improving
performance on complex mathematical and programming tasks
\cite{snell2024scaling, guo2025deepseek}. Multilingual reasoning extends
this capability across languages, requiring models to understand a problem
and carry out its inferential steps beyond English. Existing studies show
that LLMs can reason across many languages, but that their performance and
reasoning behavior vary substantially across linguistic settings
\cite{shilanguage, wang2025}.

To understand these cross-lingual differences, a growing line of work investigates multilingual reasoning through reasoning language, the language in which a model expresses its intermediate reasoning. Motivated by the observation that reasoning models often default to English even when prompted in another language, these studies manipulate reasoning language at inference time, prompting or steering a fixed model to reason in the target language and examining whether this improves accuracy, cross-lingual consistency, or language compliance~\cite{shilanguage, qi-etal-2025-models, zhao2026}. This line of work has clarified how generation-time language affects the behavior of an already-trained model. Because the model parameters and training history remain fixed, however, it cannot reveal how reasoning-language behavior is acquired, or whether the same student would develop differently when trained on the same reasoning content expressed in a different linguistic form. Addressing this question requires moving the intervention from inference time to training time.

In the training-time setting, recent work has increasingly used multilingual chain-of-thought distillation, which requires choosing the linguistic form in which teacher reasoning is presented to the student. Prior work has explored English-derived target-language traces~\cite{chen2024, laiandnissim2024, barua2025long}, English-pivot supervision~\cite{zhu2024, ko2025}, and language-mixed traces~\cite{chai2025, son2025}. These studies report that different supervision recipes can lead to different downstream outcomes, but they do not fully identify why. The compared datasets are often constructed through different generation, translation, and filtering pipelines, so they may vary not only in linguistic form but also in the reasoning trajectory itself, including step structure, explanatory detail, and trace quality. A downstream gap between English, target-language, and mixed-language supervision therefore cannot by itself be attributed to linguistic form. We address this identification problem by deriving all three shells from
the same verified English reasoning trajectory, preserving its underlying
reasoning content---including operations, equations, step order,
intermediate conclusions, and the final answer---while varying its
response-side linguistic realization. This content-controlled design allows us to test whether the three stages of the
supervision lifecycle---compatibility, learnability, and inheritance---align.

We conduct a content-controlled study of multilingual mathematical reasoning by measuring pre-training compatibility, post-fine-tuning accuracy, and the linguistic form of generated reasoning traces. We find three main results even when the underlying reasoning content
is held fixed. First, pre-training likelihood yields consistent shell rankings across students
but does not reliably predict downstream learnability. Second, shell effects depend on the student, language, and difficulty, with less compatible shells outperforming more compatible alternatives in some settings. Third, output-language behavior can diverge from downstream accuracy:
shells with no clear accuracy gap in our tests can nonetheless produce
sharply different language profiles. Supervision-shell selection is therefore a setting-dependent, multi-objective decision that should be guided by the intended deployment objectives rather than by a single proxy.

Overall, we make the following contributions:
\textbf{(i)} We introduce a content-controlled language-shell framework
for multilingual reasoning supervision by rendering the same verified
reasoning trajectory into English, translated target-language, and
language-mixed shells.
\textbf{(ii)} We evaluate supervision shells across three stages:
pre-training compatibility, downstream learnability, and post-training
inheritance, using two pretrained student models, three languages, and
two difficulty levels, together with a focused layerwise case study.
\textbf{(iii)} We demonstrate that these three stages do not consistently align and derive two practical guidelines: avoid excluding candidate shells based solely on pre-training compatibility, and evaluate both downstream accuracy and output-language behavior.

\begin{figure*}[t]
  \centering
  \includegraphics[width=\textwidth]{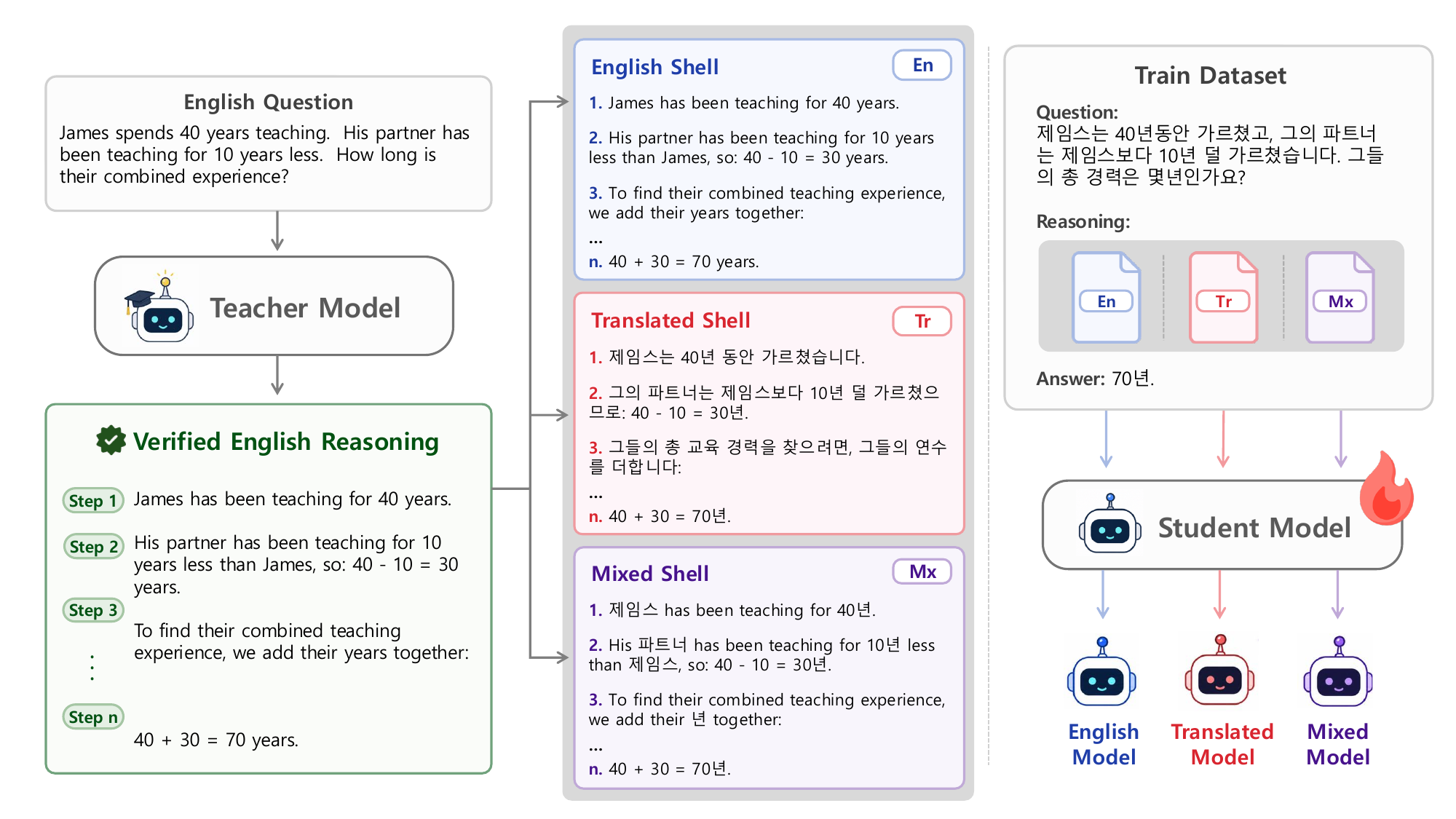}
  \caption{The content-controlled language-shell framework (Korean example). A verified
English teacher trajectory is rendered as three response-side shells:
English (original), Translated, and Mixed. Each shell is paired with the same
target-language question and answer, and a separate student initialized from
the same checkpoint is fine-tuned on the corresponding dataset, making
response-side realization the intended contrast.}
  \label{fig:main_framework}
\end{figure*}
\section{Related Work}

\subsection{Analyzing Multilingual Reasoning Behavior}

A substantial body of work has examined how language choice and cross-lingual processing shape multilingual reasoning beyond final-answer accuracy. Comparisons between English and target-language reasoning reveal persistent weaknesses in target-language long-form reasoning and potential advantages of English as a reasoning lingua franca~\cite{barua2025long, saji-etal-2026-reasoning}. Complementary analyses investigate the reasoning process itself, evaluating language compliance, cross-lingual consistency, and trace faithfulness, and tracing much of the multilingual reasoning gap to failures in understanding non-English inputs before subsequent reasoning begins~\cite{zhao2026, kang2026}. Together, these studies show that multilingual reasoning depends not only on final-answer generation but also on reasoning language, trace reliability, and input understanding. However, these studies characterize models whose training is already
complete, leaving unclear how the language form of supervision itself
shapes what the same student acquires. We address this gap by holding
reasoning content fixed and varying its response-side language shell.

\subsection{Multilingual Reasoning Supervision and Distillation}

Prior work has improved multilingual reasoning through several forms of
supervision: English-centered approaches that align non-English questions
with an English reasoning space~\cite{zhu2024,tran2026reasoning};
target-language supervision, including corpora translated from English
traces~\cite{laiandnissim2024,chen2024}, which can outperform newly
distilled target-language traces and, under matched compute, English-only
supervision in some settings~\cite{barua2025long}; and mixed-language
supervision combining English scaffolds with target-language
elements~\cite{chai2025,son2025}. Progressive Code-Switching further uses
step-level mixing as an intermediate curriculum toward target-language
reasoning~\cite{wang2026efficientmultilingualreasoningtransfer}. However,
prior work generally does not construct English, translated, and mixed
shell conditions as paired realizations of the same verified trajectory,
leaving language form entangled with generation, translation, filtering,
or mixing procedures. We address this by deriving all three shells from
the same teacher trajectory and fine-tuning a separate student
per shell from the same initialization.

\begin{figure}[t]
  \centering 
  \includegraphics[width=\columnwidth]{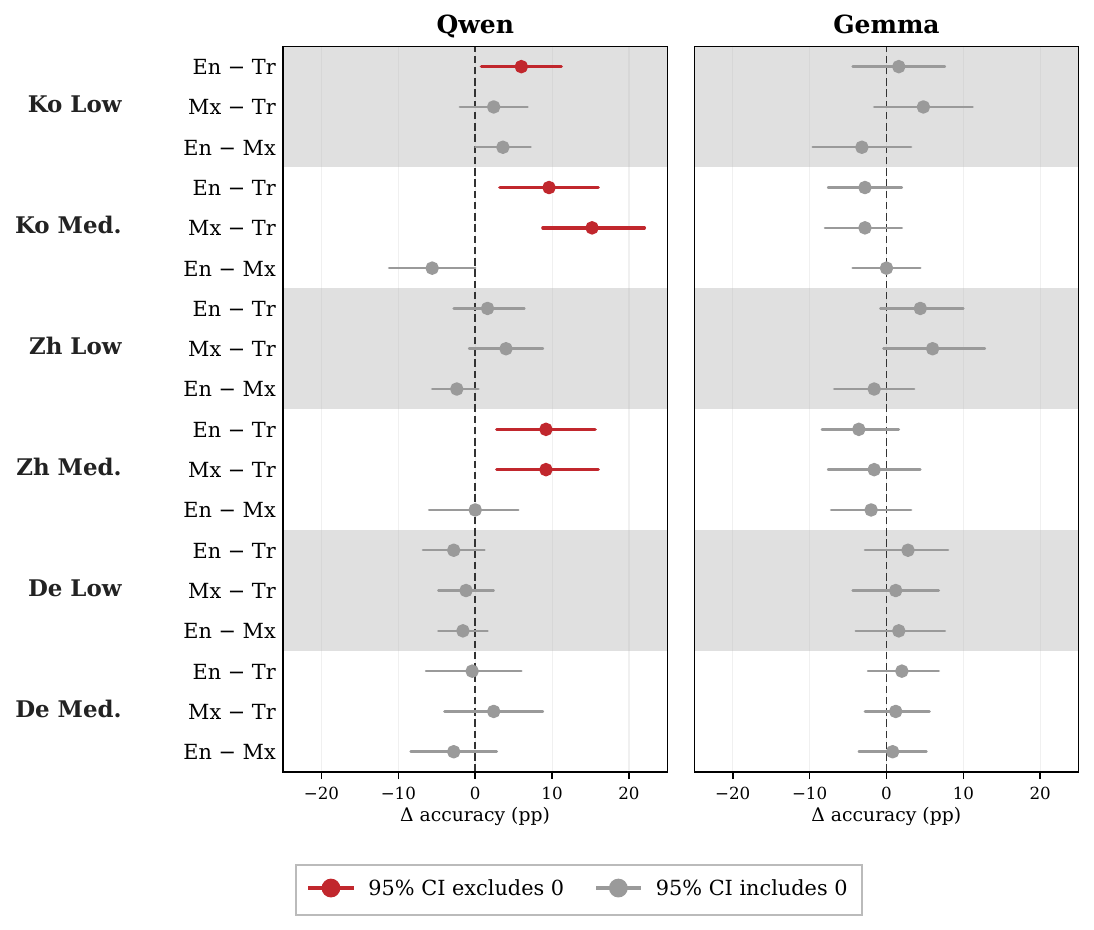}
  \caption{Pairwise shell effects with seed-averaged, item-level
    paired-bootstrap nominal 95\% confidence intervals; positive values favor
    the first shell. En, Tr, and Mx denote the English, Translated, and Mixed
    shells, respectively. All five intervals excluding zero occur for Qwen,
    with English or Mixed achieving higher accuracy than Translated.
    Ko/Zh/De denote Korean/Chinese/German, and Low/Med.\ denote the PolyMath
    difficulty levels.}
  \label{fig:pairwise_shell_effects}
\end{figure}
\section{Controlled Language-Shell Analysis Framework}
\label{sec:framework}

We formulate our study as a controlled language-shell comparison that
holds the reasoning content fixed while varying its response-side
linguistic realization. As illustrated in
Figure~\ref{fig:main_framework}, the framework keeps the student-side
target-language question and underlying reasoning content identical
across conditions, making the response-side realization the intended
contrast. This design enables a direct comparison of supervision shells, with the
intended reasoning steps, intermediate conclusions, explanatory detail,
and solution quality matched across conditions.
Section~\ref{sec:shell_definition} formalizes the distinction between
reasoning content and language shell and describes the controlled
variants. Section~\ref{sec:lifecycle_dimensions} then defines the three
lifecycle dimensions used to compare them: compatibility, learnability,
and shell inheritance.

\subsection{Separating Reasoning Content from Language Shell}
\label{sec:shell_definition}

For each problem $i$, let $x_i^{\mathrm{en}}$ denote the English version presented to the teacher and $x_i^{\ell}$ the corresponding version in target language $\ell$ presented to the student. A single English teacher response whose final answer is verified as correct serves as the reference trajectory. We denote its intended reasoning content by $c_i$. All shell-specific training examples pair the same student-side question $x_i^{\ell}$ with alternative linguistic realizations of this reference trajectory.

We define \emph{reasoning content} as the ordered inferential structure of a
solution, including the relevant quantities and relations, operations,
intermediate conclusions, step ordering, and final answer. Let
\(\mathcal{S} =
\{\textsc{English}, \textsc{Translated}, \textsc{Mixed}\}\) denote the set of
language shells. A \emph{language shell} \(s \in \mathcal{S}\) specifies how
this content is linguistically realized, allowing surface properties such as
lexical choice, syntax, script, and the language of individual segments to
vary. Let \(R_s\) denote the rendering procedure associated with shell \(s\),
such that
\begin{equation}
    y_i^{s} = R_s(c_i),
\end{equation}
where \(y_i^{s}\) is the shell-specific supervision response for problem \(i\).

The \textbf{English} shell retains the original teacher response. The \textbf{Translated} shell translates the explanatory text of the reference trajectory into the target language while preserving its ordered reasoning structure, rather than independently generating a new target-language solution. The \textbf{Mixed} shell retains most of the logical reasoning scaffolding in English while rendering problem-specific key terms in the target language, an entity-level localization rather than sentence-level mixing, following the Language-Mixed CoT schema of~\citet{son2025}.
All three shells are therefore alternative renderings of the same English-derived trajectory; independently generated target-language solutions are excluded to preserve content control.

By rendering the same reference trajectory in three response-side forms, we
treat language shell as the primary varying factor while holding the
underlying reasoning content fixed. We evaluate content preservation rather
than assuming it, as described in Appendix~\ref{app:audit_results}. Shell
rendering also changes correlated surface properties, including sequence
length, tokenization, and the rendering model itself. We therefore treat
each shell as a complete response-side realization rather than isolating
its individual components, and examine sequence length
(Section~\ref{sec:length_analysis}) and renderer identity
(Appendix~\ref{app:renderer_control}) as alternative explanations.

\subsection{Compatibility, Learnability, and Shell Inheritance}
\label{sec:lifecycle_dimensions}

We examine language-shell supervision at three stages of the supervision lifecycle: compatibility before fine-tuning, learnability through shell-specific fine-tuning, and inheritance after fine-tuning. 

\paragraph{Compatibility.}
Compatibility measures how probable each shell-specific response is under the initial student before fine-tuning. For problem $i$ and shell $s$, we compute the token-normalized conditional log-likelihood of $y_i^{s}$ given the target-language question $x_i^{\ell}$:
\begin{equation}
    C_{i,s}
    =
    \frac{1}{|y_i^{s}|}
    \sum_{t=1}^{|y_i^{s}|}
    \log
    \pi_0
    \left(
        y_{i,t}^{s}
        \mid
        x_i^{\ell},
        y_{i,<t}^{s}
    \right),
\end{equation}
We report compatibility as the corresponding perplexity,
\begin{equation}
    \mathrm{PPL}_{i,s}
    =
    \exp\left(-C_{i,s}\right),
\end{equation}
where lower values indicate greater initial compatibility. For each shell,
we average \(C_{i,s}\) across items and exponentiate the negative mean to
obtain the reported perplexity. Following the likelihood-based
response-selection approach of \citet{zhang2025}, we use this quantity as
an operational measure of pre-training compatibility.

\paragraph{Learnability.}
Learnability measures downstream reasoning performance after fine-tuning on
each shell. We operationalize it as post-training final-answer accuracy under
a fixed evaluation protocol. Within each model--language condition, all
shell-trained students share the same initialization, content-matched
training items, training configuration, and evaluation protocol. We therefore
use learnability to denote the total downstream outcome associated with each
training shell under this shared inference setting, including effects
potentially mediated by the language adopted during generation, rather than
a language-invariant measure of reasoning ability.

\paragraph{Shell Inheritance.}
Shell inheritance measures how shell-specific fine-tuning is reflected in the
trained student's outputs. Our primary measure, \emph{behavioral inheritance},
captures the language composition of generated reasoning after shell-specific
fine-tuning, quantified by target-language proportion. As a focused case
study, we additionally examine layerwise Korean-token accessibility for Qwen.
Neither analysis directly identifies the language in which the model
internally reasons.

Together, these measurements compare each shell's pre-training fit,
downstream accuracy, and generated output-language composition.

\begin{table}[t]
\centering
\scriptsize
\setlength{\tabcolsep}{1.8pt}
\renewcommand{\arraystretch}{1.05}
\begin{tabular*}{\columnwidth}{
    @{\extracolsep{\fill}}
    lrrrrrr
    @{}
}
\toprule
&
\multicolumn{3}{c}{Qwen} &
\multicolumn{3}{c}{Gemma} \\
\cmidrule(lr){2-4}
\cmidrule(lr){5-7}
\raisebox{2.2ex}[0pt][0pt]{Shell}
& Ko & Zh & De
& Ko & Zh & De \\
\midrule
English
& \textbf{2.08} & \textbf{2.03} & 1.99
& \textbf{2.21} & \textbf{2.24} & \textbf{2.20} \\
Translated
& 2.33 & 2.18 & \textbf{1.98}
& 2.23 & 2.41 & 2.21 \\
Mixed
& 2.55 & 2.11 & 2.07
& 2.28 & 2.32 & 2.26 \\
\bottomrule
\end{tabular*}
\caption{
Pre-training compatibility, reported as the perplexity of each shell's
supervision responses under the initial student using the training prompt
template. Lower values indicate greater compatibility, and the lowest value
within each student--language condition is shown in \textbf{bold}.
}
\label{tab:shell_compatibility}
\end{table}

\section{Experimental Setup}
\label{sec:setup}

\subsection{Data Construction and Validation}
\label{sec:data_construction}

We use Qwen3-32B~\cite{yang2025} to generate an English reference
trajectory for each problem in the GSM8K training
set~\cite{cobbe2021}, using a fixed GSM8K-style system prompt and
retaining only trajectories whose extracted final answers match the
gold answers. Target-language questions are
obtained by translating each GSM8K question with
Gemini-2.5-Flash~\cite{comanici2025}. We then render each verified
trajectory into the three shells defined in Section~\ref{sec:shell_definition},
with Gemini-2.5-Flash producing the Translated and Mixed
renderings. For each target language, we retain only items whose responses
under all three shell conditions yield extractable final answers matching the
gold answer. The resulting datasets contain approximately 7.2K shared items
per target language. Consequently, the English, Translated, and Mixed datasets
within each language have identical item IDs and differ only in their
response-side linguistic rendering.

We validate content preservation using a GPT-4o judge~\cite{hurst2024gpt4o} with shell-specific rubrics. The judge is applied to all pairs whose equation sets have a Jaccard similarity below 0.9, together with a random sample of the remaining pairs. Mean scores exceed 4.7 out of 5 for every preservation dimension
(most exceed 4.9), and the judge identifies no added or omitted
information in at least 98.8\% of evaluated pairs. Further details on data construction and validation are provided in Appendix~\ref{app:data_construction}.

\subsection{Models and Training}
\label{sec:models_training}

Our student backbones are Qwen3-4B-Base~\cite{yang2025} and
Gemma-3-4B-PT~\cite{kamath2025}. We use pretrained rather than
instruction-tuned checkpoints to examine shell-specific learning without
introducing prior instruction-alignment effects. We perform full-parameter
supervised fine-tuning with loss applied only to response tokens. Every
shell condition within a student--language pair is trained with identical
data size, step count, and optimization settings. A fixed
language-specific prompt template is shared across training and all
measurements. We train two random seeds per condition and report
seed-averaged results. Appendix~\ref{app:training_details} provides
further training details.

\subsection{Evaluation}
\label{sec:evaluation}

For \textbf{Compatibility}, we compute the initial student's
token-normalized conditional log-likelihood of each shell response under the
training prompt template and report the corresponding perplexity.

For \textbf{Learnability}, we evaluate final-answer accuracy on the Low and
Medium subsets (125 problems each) of PolyMath~\cite{wang2025} in the
corresponding target language for each model--language condition. Generation
uses the language-specific prompt template shared with training, greedy
decoding, language-specific answer markers, and budgets of 2,048 and 4,096
new tokens for Low and Medium, respectively. Answers are extracted and
scored with a deterministic rule-based checker applied identically to all
shell conditions. For each item, we first average outcomes across seeds and
then estimate pairwise shell differences using item-level paired-bootstrap
95\% confidence intervals. Intervals are nominal, without correction for
multiple comparisons.

For \textbf{Shell Inheritance}, we measure the target-language proportion of
each generated reasoning trace over its prose, excluding equations, digits,
and symbols. We use character-level proportions for Korean and Chinese and
word-level proportions for German, and compare them across shell conditions
and difficulty levels. Further evaluation details are provided in
Appendix~\ref{app:training_details}.

\begin{table*}[t]
\centering
\small
\begin{tabular*}{\textwidth}{
    @{\extracolsep{\fill}}
    ll
    cc
    cc
    cc
    @{}
}
\toprule
\multirow{2}{*}{\raisebox{-0.6ex}[0pt][0pt]{Model}}
& \multirow{2}{*}{\raisebox{-0.6ex}[0pt][0pt]{Shell}}
& \multicolumn{2}{c}{Ko}
& \multicolumn{2}{c}{Zh}
& \multicolumn{2}{c}{De} \\
\cmidrule(lr){3-4}
\cmidrule(lr){5-6}
\cmidrule(lr){7-8}
&
& Low & Med.
& Low & Med.
& Low & Med. \\
\midrule
\multirow{4}{*}{Qwen}
& Base
& 64.8 & 19.2
& 72.0 & 18.4
& 36.0 & 8.0 \\
& English
& \textbf{88.8} & \underline{35.2}
& \underline{84.0} & \textbf{44.4}
& 81.2 & 32.0 \\
& Translated
& 82.8 & 25.6
& 82.4 & \underline{35.2}
& \textbf{84.0} & \underline{32.4} \\
& Mixed
& \underline{85.2} & \textbf{40.8}
& \textbf{86.4} & \textbf{44.4}
& \underline{82.8} & \textbf{34.8} \\
\midrule
\multirow{4}{*}{Gemma}
& Base
&  4.0 &  6.4
&  8.0 &  6.4
&  8.0 &  8.0 \\
& English
& \underline{59.2} & \underline{11.2}
& \underline{60.8} &  9.6
& \textbf{62.0} & \textbf{11.6} \\
& Translated
& 57.6 & \textbf{14.0}
& 56.4 & \textbf{13.2}
& 59.2 &  9.6 \\
& Mixed
& \textbf{62.4} & \underline{11.2}
& \textbf{62.4} & \underline{11.6}
& \underline{60.4} & \underline{10.8} \\
\bottomrule
\end{tabular*}
\caption{Downstream final-answer accuracy (\%) on the Low and Medium subsets of
PolyMath after shell-specific fine-tuning. Base denotes the initial
checkpoint. The highest value within each student--language--difficulty
condition is shown in \textbf{bold} and the second-highest is
\underline{underlined}.}
\label{tab:polymath_main}
\end{table*}

\begin{table}[t]
\centering
\scriptsize
\setlength{\tabcolsep}{3.0pt}
\renewcommand{\arraystretch}{1.05}
\begin{tabular*}{\columnwidth}{
    @{\extracolsep{\fill}}
    llccc
    @{}
}
\toprule
\raisebox{-1.7ex}[0pt][0pt]{Student}
&
\raisebox{-1.7ex}[0pt][0pt]{Shell}
& Ko
& Zh
& De \\
\cmidrule(lr){3-3}
\cmidrule(lr){4-4}
\cmidrule(lr){5-5}
&
& Low $\rightarrow$ Med.
& Low $\rightarrow$ Med.
& Low $\rightarrow$ Med. \\
\midrule
\multirow{4}{*}{Qwen}
& Base       & 98.0 $\rightarrow$ 97.4 & 98.8 $\rightarrow$ 98.3 & 70.0 $\rightarrow$ 75.9 \\
& English    &  0.8 $\rightarrow$  0.3 &  0.7 $\rightarrow$  0.3 &  0.1 $\rightarrow$  0.0 \\
& Translated & 98.6 $\rightarrow$ 99.2 & 99.2 $\rightarrow$ 99.4 & 76.8 $\rightarrow$ 74.7 \\
& Mixed      & 12.6 $\rightarrow$  1.0 & 13.1 $\rightarrow$  1.0 & 10.1 $\rightarrow$  1.0 \\
\midrule
\multirow{4}{*}{Gemma}
& Base       & 99.1 $\rightarrow$ 94.2 & 99.2 $\rightarrow$ 95.5 & 67.0 $\rightarrow$ 35.7 \\
& English    &  0.7 $\rightarrow$  0.3 &  0.7 $\rightarrow$  0.3 &  0.0 $\rightarrow$  0.0 \\
& Translated & 98.8 $\rightarrow$ 99.6 & 99.1 $\rightarrow$ 99.3 & 76.8 $\rightarrow$ 76.4 \\
& Mixed      & 12.3 $\rightarrow$  1.9 & 13.0 $\rightarrow$  2.3 & 10.1 $\rightarrow$  1.5 \\
\bottomrule
\end{tabular*}
\caption{
Target-language proportion (\%) of generated reasoning by shell, language,
and difficulty level. Base denotes the initial checkpoint before
fine-tuning.
}
\label{tab:language_inheritance}
\end{table}
\section{Results}

Having fixed the reasoning trajectory across shells, we examine three
main patterns across the supervision lifecycle: pre-training compatibility
does not reliably track learnability
(\S\ref{sec:compatibility}); learnability varies across students,
languages, and difficulty levels rather than following a universal shell
ordering (\S\ref{sec:learnability}); and output-language behavior can
diverge from accuracy: English and Translated induce sharply different
profiles, while Mixed becomes more English-dominant on Medium
(\S\ref{sec:inheritance}).

\subsection{Compatibility}
\label{sec:compatibility}

\paragraph{Compatibility rankings are consistent across students.}
As shown in Table~\ref{tab:shell_compatibility}, both students exhibit largely the same
language-specific compatibility ordering. In Korean, both rank the shells as
English $>$ Translated $>$ Mixed. In Chinese, both rank them as
English $>$ Mixed $>$ Translated. In German, English and Translated are
nearly tied, while Mixed is the least compatible shell. Thus, token-normalized compatibility rankings are largely consistent across students: English generally receives the lowest perplexity, whereas Mixed is never ranked first.

\paragraph{Compatibility does not reliably track learnability.}
This consistent compatibility ordering, however, does not translate into a
corresponding ordering of downstream learnability. As shown in
Table~\ref{tab:polymath_main}, Mixed is numerically strongest or tied for
strongest in 6 of the 12 PolyMath settings. The clearest reversal occurs for
Qwen on Korean Medium: its compatibility ordering is
English $>$ Translated $>$ Mixed, whereas its downstream accuracy follows
Mixed $>$ English $>$ Translated. Mixed exceeds Translated by 15.2 percentage
points in this setting.

\subsection{Learnability}
\label{sec:learnability}

\paragraph{Content-matched shells need not yield equivalent learning outcomes.}

Table~\ref{tab:polymath_main} reports downstream accuracy after fine-tuning
each student on the three content-matched shells. All shell-trained models
improve over their corresponding base checkpoints, indicating
a strong common benefit from the shared reasoning supervision. However, the
same trajectory is not learned equally across its linguistic renderings. As
shown in Figure~\ref{fig:pairwise_shell_effects}, five pairwise contrasts have
nominal 95\% confidence intervals excluding zero, all concentrated in Qwen
and all disfavoring Translated relative to English or Mixed, with gaps ranging
from 6.0 to 15.2 percentage points. No such interval excludes zero in any German condition or for Gemma, though Gemma's near-floor Medium accuracy may limit the power to detect shell effects. Thus, even when the underlying reasoning
content is matched, different linguistic renderings can lead to different
learning outcomes. These differences are concentrated in particular students,
languages, and difficulty levels rather than following a universal shell
ordering.

\paragraph{Mixed remains competitive despite low pre-training compatibility.}

As noted above, Mixed is numerically strongest or tied for strongest in 6 of the 12
downstream settings, although most corresponding pairwise confidence
intervals include zero. More specifically, no English--Mixed contrast has a
nominal 95\% confidence interval excluding zero, whereas Mixed exceeds
Translated in Qwen's Korean and Chinese Medium settings
(Figure~\ref{fig:pairwise_shell_effects}). Rather than establishing Mixed
as universally superior, these results show that low initial compatibility
does not preclude competitive downstream learning: a lower-ranked shell
under this metric can remain competitive with or, in the Qwen Korean and
Chinese Medium settings, outperform higher-ranked alternatives. Taken
together, these results show that shell choice can modulate learning even
when reasoning content is controlled, while no single shell dominates
across students and conditions. We further examine the robustness of these findings through additional-seed experiments and cross-benchmark evaluation on MATH-500 (Appendices~\ref{app:training_run_variability} and~\ref{app:math500}, respectively).

\begin{figure}[t]
  \centering
  \includegraphics[width=1\columnwidth]{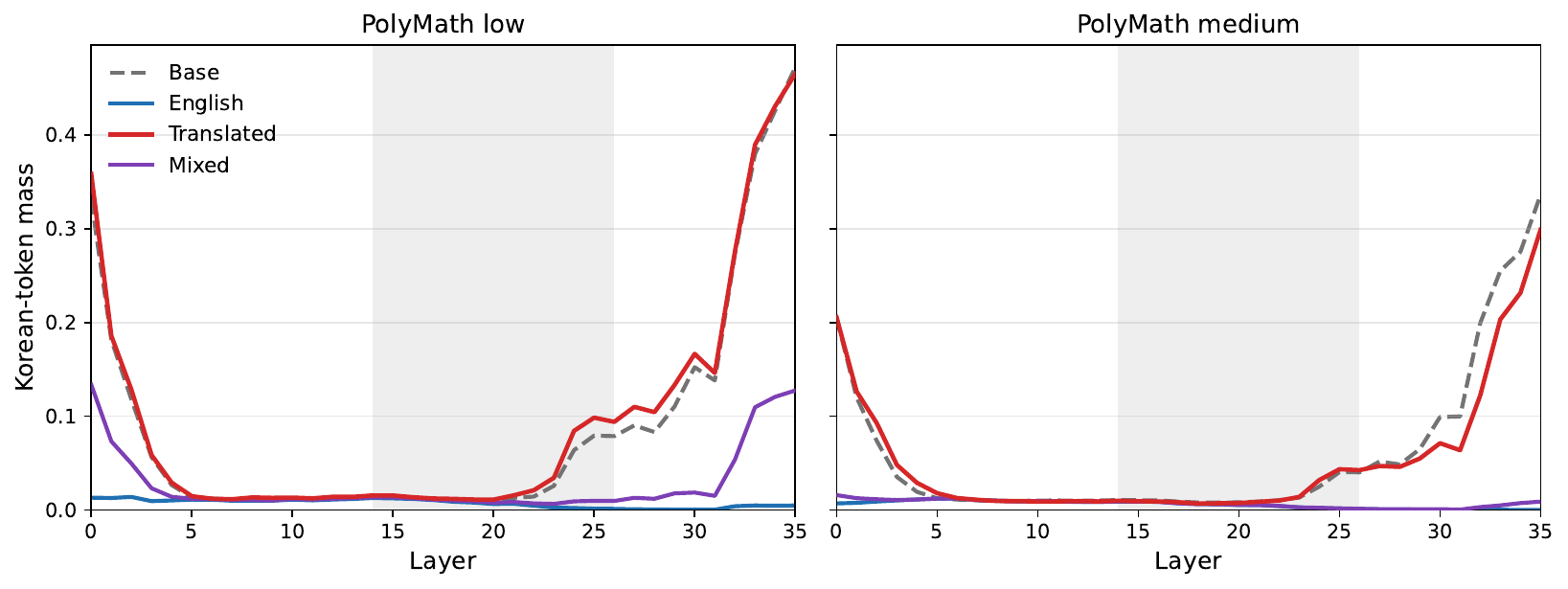}
    \caption{
Layerwise Korean-token probability mass over all generated positions on the
Low (left) and Medium (right) subsets of PolyMath, obtained by projecting
intermediate hidden states of each Korean-condition Qwen student (Base and
the three shell-trained models) through the output head.
    }
  \label{fig:logit_lens}
\end{figure}

\begin{table}[t]
\centering
\scriptsize
\setlength{\tabcolsep}{4pt}
\renewcommand{\arraystretch}{1.05}

\begin{tabular}{llrr}
\toprule
\multicolumn{4}{l}{\textbf{A. Supervision length ratio (Translated/English)}} \\
Student & Ko & Zh & De \\
\midrule
Qwen     & 1.31 & 1.06 & 1.23 \\
Gemma & 1.15 & 1.04 & 1.11 \\
\midrule
\multicolumn{4}{l}{\textbf{B. Generated-length regressions (Qwen, Medium)}} \\
Lang.\ / Subset
& $\Delta$Acc
& \multicolumn{2}{r}{$\beta$ for Tr./En.\ length ($p$)} \\
\midrule
Ko / All   & $+9.6$  & \multicolumn{2}{r}{$+0.058$ (.16)} \\
Ko / Comp. & $+13.3$ & \multicolumn{2}{r}{$+0.096$ (.35)} \\
Zh / All   & $+9.2$  & \multicolumn{2}{r}{$-0.027$ (.43)} \\
Zh / Comp. & $+9.5$  & \multicolumn{2}{r}{$-0.349$ (.042)} \\
\bottomrule
\end{tabular}

\caption{Sequence-length analysis for the English--Translated contrast
($\Delta$Acc = English $-$ Translated, in pp). Panel A reports
Translated-to-English supervision-length ratios under each student
tokenizer. Panel B reports slopes ($\beta$, with $p$-values) from
regressing the item-level paired correctness difference on the
Translated-to-English generated-length ratio; Comp.\ restricts the
analysis to items for which both models complete generation within the
token budget.}
\label{tab:length_analysis}
\end{table}

\subsection{Shell Inheritance}
\label{sec:inheritance}

\paragraph{English and Translated shells induce sharply different
output-language behavior.}
Table~\ref{tab:language_inheritance} shows that shell-specific fine-tuning
strongly affects the language of generated reasoning. English-trained
students produce almost no target-language output (under 1\%), whereas
Translated-trained students generate predominantly in the target language
(over 70\%). This separation is consistent across both students and all
three languages, and remains largely stable across difficulty levels.
This sharp separation in output language also appears in conditions where
the corresponding accuracy contrast is not clearly resolved by our tests.
Thus, downstream accuracy alone does not fully characterize shell
inheritance.

\paragraph{Mixed outputs become consistently more English-dominant on Medium.}
Mixed supervision exhibits a distinct cross-subset pattern. Across both
students and all three languages, Mixed outputs contain a modest
target-language proportion on Low (10--13\%), which falls to roughly
1--2\% on Medium. By contrast, the output-language composition induced
by English and Translated remains comparatively stable. Thus, Mixed
shows a consistent shift toward predominantly English output across all
six model--language comparisons. Because target-language proportion
depends on output composition and the availability of localizable terms,
this pattern does not by itself establish reduced adherence to the full
Mixed schema or a causal effect of difficulty.
Section~\ref{sec:logitlens} examines whether this cross-subset behavioral
shift is accompanied by corresponding changes in intermediate token
accessibility.

\section{Analysis}

\subsection{Are Shell Effects Explained by Sequence Length?}
\label{sec:length_analysis}

Because translation into Korean, Chinese, or German systematically lengthens
the supervision traces under the student tokenizers, sequence length is a
natural alternative explanation for the English--Translated performance
differences. We therefore examine length at both the supervision and
generation stages. Table~\ref{tab:length_analysis} reports the key
English--Translated results, with additional comparisons against both English
and Mixed provided in Appendix~\ref{app:sequence_length}.

\paragraph{Supervision-side length analysis.}
To test whether longer supervision sequences account for the observed
Translated-shell disadvantages, we compare content-matched English and
Translated supervision lengths. Panel~A of Table~\ref{tab:length_analysis}
confirms that Translated traces are consistently longer under both student
tokenizers. However, the ordering of these ratios does not match the
downstream pattern in Section~\ref{sec:learnability}: German has among the
largest ratios but no confidence interval excluding zero, whereas Chinese has
the smallest ratio despite a clear English--Translated gap for Qwen on Medium
problems. Supervision length therefore does not consistently explain the
observed differences.

\paragraph{Generated-response length analysis.}
To test whether generated-response length accounts for the observed accuracy
gaps, we regress the item-level paired correctness difference on the
Translated-to-English generated-length ratio in the Qwen Medium settings. As
shown in Panel~B of Table~\ref{tab:length_analysis}, neither slope is
distinguishable from zero for the full Korean and Chinese subsets, and
restricting the analysis to responses completed within the generation budget
leaves the accuracy gaps unchanged or larger. The only nominal association,
in the completed Chinese subset, is negative: greater relative Translated
length is associated with a smaller, rather than larger, English advantage.

Together, these analyses show that the Translated shell is systematically
longer than the English shell, but this length difference does not align with
its downstream accuracy disadvantage. Because length varies together with
other shell properties, however, these analyses cannot establish whether
length itself causes any of the observed differences.

\subsection{Layerwise Representational Case Study}
\label{sec:logitlens}

As a focused case study, we examine whether the lower target-language
proportion of Mixed outputs on Medium observed in
Section~\ref{sec:inheritance} is accompanied by changes in intermediate
token accessibility. Using a logit lens, we measure Korean-token
probability mass across layers of the Korean Qwen models on Low and
Medium. As shown in Figure~\ref{fig:logit_lens}, Base, English, and
Translated show similar profiles across subsets, whereas Mixed retains a
visible late-layer Korean component on Low that is largely absent on
Medium, leaving its Medium profile close to English. Thus, on Medium,
Mixed produces more English-dominant output while its late layers assign
lower probability mass to Korean tokens, bringing its profile closer to
that of the English-trained model, without clear mid-layer differences.
This parallel between generated output and late-layer token accessibility
is consistent with prior observations linking generated and hidden-layer
script composition \citep{wang-etal-2025-language-mixing}, but does not
identify a shared mechanism or internal reasoning language.

\section{Discussion}

\subsection{Implications for Multilingual Reasoning Supervision}

Our findings suggest two practical guidelines for supervision-shell selection. 

First, pre-training compatibility should not be treated as a standalone proxy for learnability. For Qwen in Korean, Mixed has the lowest initial compatibility (Table~\ref{tab:shell_compatibility}) yet achieves the highest accuracy on PolyMath Medium (Table~\ref{tab:polymath_main}). Thus, low compatibility alone should not justify excluding a candidate shell.

Second, shell selection should consider both accuracy and output-language requirements. For Qwen in German on Medium, English and Translated achieve similar accuracy (32.0\% vs.\ 32.4\%) but substantially different target-language proportions (0.0\% vs.\ 74.7\%), demonstrating that accuracy alone does not capture shell inheritance.

Together, these findings establish shell selection as a setting-dependent, multi-objective decision. Our content-controlled framework and three-stage analysis provide a basis for developing and evaluating better supervision-selection signals.

\subsection{Accuracy and Output Language Can Diverge under Mixed Supervision}

Language-mixed supervision combines target-language terms with an
English reasoning scaffold \citep{son2025}. Yet for Qwen on the
Medium-difficulty Korean and Chinese subsets, Mixed outperforms Translated
even though its generated reasoning contains only roughly 1--2\%
target-language content (\S\ref{sec:inheritance}). Thus, Mixed can retain an accuracy advantage despite reverting to
predominantly English generation, showing that the supervision shell can shape downstream accuracy and
generated reasoning language separately.
\section{Conclusion}

We studied multilingual reasoning supervision under content control,
rendering the same verified English trajectories into English, translated
target-language, and language-mixed shells. Our results show that the
language shell is a meaningful design factor in supervision: even with the
reasoning content held fixed, changing the response-side shell can alter
downstream accuracy, with effects that depend on the student, language, and
difficulty. Neither compatibility nor accuracy alone suffices for shell
selection: pre-training compatibility does not reliably track downstream
accuracy, and accuracy does not fully characterize output-language behavior.
Notably, for Qwen on the Medium-difficulty Korean and Chinese subsets, Mixed
outperforms Translated despite producing predominantly English reasoning.
Supervision-shell selection is thus a setting-dependent, multi-objective
decision, best guided by the intended deployment objectives.
\section*{Limitations}

Our controlled design enables a focused comparison of response-side shells,
but it also narrows the scope of the conclusions and constrains the scale of
the experimental grid. We discuss these limitations below.

\paragraph{Scope of the controlled contrast.}
All three shells are rendered from the same English teacher trajectory, so our study isolates the response-side realization of English-derived reasoning rather than natively conceived target-language reasoning. Reliance on proprietary models limits reproducibility, and the human evaluation does not cover all languages or question translations. Moreover, translationese and English-origin explanatory conventions may persist in the Translated shell, while the English scaffold and localized elements of the Mixed shell remain entangled.

\paragraph{Scale and statistical resolution.}
Full-parameter fine-tuning constrained our experimental grid to two pretrained students, three languages, and two difficulty levels. While the main analysis uses two seeds per condition, we additionally examine training-run variability using four seeds for Qwen in Korean and Chinese (Appendix~\ref{app:training_run_variability}). However, the remaining conditions retain two seeds, limiting statistical resolution. Gemma's near-floor Medium accuracy further limits our ability to detect shell effects. Broader model, language, and task coverage, together with additional seeds, is needed to establish the generality of our findings.

\section*{Acknowledgments}
This work was supported by the Institute of Information \& Communications Technology Planning \& Evaluation (IITP) grant funded by the Ministry of Science and ICT (MSIT), Korea (RS-2021-II211341, Artificial Intelligence Graduate School Program, Chung-Ang University), and by the National Research Foundation of Korea (NRF) grants funded by MSIT (RS-2025-00556246 and RS-2021-NR057239).


\bibliography{custom}

\clearpage

\appendix

\label{sec:appendix}

\section{Data Construction and Validation}
\label{app:data_construction}

\subsection{English Teacher Trajectory Generation}
\label{app:teacher_generation}

We generate one English reference trajectory for each GSM8K training
problem using Qwen3-32B. The English question is supplied as the user
input, together with the following fixed system prompt:

\begin{quote}
\small\ttfamily
You are an expert math tutor. Solve the given grade-school math word
problem step by step.\\
End with exactly this final line: The answer is N.
\end{quote}

\subsection{Question Translation and Dataset Sizes}
\label{app:question_translation}

We begin with all \textbf{7,473} problems in the GSM8K training split.
Target-language questions are obtained by translating each question into
the target language with Gemini-2.5-Flash, using the system prompt in
Table~\ref{tab:question_translation_prompt} with the English question
supplied as the user input. For each target
language, we retain only items whose responses under all three shell
conditions yield an extractable final answer matching the gold answer. This
filtering removes both incorrect teacher trajectories and items whose final
answer is altered during rendering. We retain \textbf{7,215} problems for Korean,
\textbf{7,167} for Chinese, and \textbf{7,204} for German. Because the Translated and Mixed
renderings are produced separately for each target language, the discarded
items differ across languages; the intersection of the three
language-specific sets contains 7,143 problems. Within each target language,
all three shell conditions have identical cardinality and item IDs: \textbf{7,215}
examples per shell for Korean, \textbf{7,167} for Chinese, and \textbf{7,204} for German.

\subsection{Shell-Rendering Prompts}
\label{app:rendering_prompts}

Tables~\ref{tab:translation_prompt} and~\ref{tab:mixed_prompt} reproduce the
system prompts used to construct the Translated and Mixed shells. The
verified English trajectory to be transformed was supplied separately as the
user input. Both prompts frame rendering as a transformation rather than a
new problem-solving task, explicitly requiring the original reasoning steps,
calculations, equations, and ordering to remain unchanged.

The Translated prompt renders all explanatory prose in the target language
while preserving mathematical and formatting elements verbatim. The Mixed
prompt retains the reasoning scaffold and sentence structure in English,
embedding target-language expressions only as noun phrases for person names
and problem-specific object terms drawn from the corresponding target-language
problem. Here, \texttt{\{lang\}} and \texttt{\{LANG\}} denote the target
language, \texttt{\{q\_ref\}} denotes the target-language problem text, and
\texttt{\{marker\}} or \texttt{\{marker\_rule\}} denotes the
language-specific final-answer format. For the Mixed prompt,
\texttt{\{SOURCE\_EXAMPLE\}} and \texttt{\{cfg['example']\}} are replaced
with a language-specific source--rewrite example. The symbol \(N\) denotes
the verified final answer.
\begin{table}[t]
\centering
\scriptsize
\renewcommand{\arraystretch}{1.05}
\begin{tabular}{@{}p{\columnwidth}@{}}
\specialrule{1pt}{0pt}{0pt}
\rowcolor{gray!18}[0pt][0pt]
\textbf{System Prompt} \\
\hline
\hline
\addlinespace[2pt]
You are a professional translator. \\
Translate the following grade-school math word problem from English into
\{lang\}. \\[3pt]
\textbf{Rules:} \\
1. Translate ONLY the natural-language text. \\
2. Keep ALL numbers, units, currency symbols, and mathematical expressions
EXACTLY as they are---do not localize, convert, or alter them (e.g.,
\texttt{\$12} stays \texttt{\$12}, not 12\begin{CJK}{UTF8}{mj}달러\end{CJK};
50 minutes becomes 50\begin{CJK}{UTF8}{mj}분\end{CJK}). \\
3. Transliterate person names naturally into \{lang\}, not the English
spelling (e.g., for Korean: Natalia $\rightarrow$
\begin{CJK}{UTF8}{mj}나탈리아\end{CJK}, Weng $\rightarrow$
\begin{CJK}{UTF8}{mj}웽\end{CJK}). \\
4. Preserve the meaning precisely so the numeric answer stays the same. \\
5. Output ONLY the translation, with no preamble, notes, or quotes. \\
\bottomrule
\end{tabular}
\caption{System prompt used to translate GSM8K questions into each target
language.}
\label{tab:question_translation_prompt}
\end{table}
\begin{table}[t]
\centering
\scriptsize
\renewcommand{\arraystretch}{1.05}
\begin{tabular}{@{}p{\columnwidth}@{}}
\specialrule{1pt}{0pt}{0pt}
\rowcolor{gray!18}[0pt][0pt]
\textbf{System Prompt} \\
\hline
\hline
\addlinespace[2pt]

You are a professional translator. \\
Translate the following math solution from English into natural
\{lang\}. \\[3pt]

\textbf{Rules:} \\

1. This is TRANSLATION, not solving---reproduce the same reasoning steps
in the same order; never re-solve, correct, or add steps. \\

2. Keep ALL numbers, calculations, LaTeX (\texttt{\$...\$},
\texttt{\$\$...\$\$}, \texttt{\textbackslash frac}), and markdown symbols
(\texttt{\#\#\#}, \texttt{**}, \texttt{---}) EXACTLY as they are; translate
only the prose around them. \\

3. Translate all prose into \{lang\}, leaving no English sentences.
For person names, use the transliteration in the \{lang\} problem below;
otherwise transliterate naturally. \\

4. End with exactly \texttt{'\{marker\}: N.'} and output ONLY the
translation, with no preamble. \\[6pt]

\texttt{[\{lang\} problem for name/term reference]} \\
\texttt{\{q\_ref\}} \\

\bottomrule
\end{tabular}

\caption{System prompt used to produce the Translated shell.}
\label{tab:translation_prompt}
\end{table}

\begin{table*}[t]
\centering
\begin{tabular}{p{1.0\textwidth}}
\specialrule{1pt}{0pt}{0pt}
\rowcolor{gray!18}
\textbf{System Prompt} \\
\hline
\hline
\addlinespace[2pt]

You are a text style converter. \\
Rewrite the given English math solution into a ``language-mixed'' style,
following the Language-Mixed CoT schema. \\[3pt]

\textbf{Rules:} \\

1. This is a REWRITING task, not a solving task. Do NOT re-solve, do NOT
change any reasoning step, its order, or any calculation. \\

2. Perform the logical scaffolding in ENGLISH: reasoning flow, connectives,
and explanations stay in English. \\

3. Convert to \{LANG\} only: proper names of people, and problem-specific
object terms (e.g., the items being counted, bought, or measured), using the
exact wording from the \{lang\} problem text below. \\

4. Do NOT convert generic nouns (problem, book, question, step) or verbs. Do NOT translate whole sentences or
clauses into \{lang\}---every sentence must remain grammatically ENGLISH,
with \{lang\} words embedded only as noun phrases. Even if the solution is
short, keep all reasoning sentences in English. \\

5. Preserve ALL numbers, equations, LaTeX, and markdown formatting
(\texttt{\#\#\#} headers, \texttt{**bold**}, \texttt{---}) EXACTLY as they
appear. \\

6. \{marker\_rule\}
\(\leftarrow\) ex: End with exactly \texttt{'Antwort: N.'} \\

7. Output ONLY the rewritten solution. No preamble. \\[6pt]

\textbf{Example of the target style:} \\

\quad Source: \texttt{"\{SOURCE\_EXAMPLE\}"} \\

\quad Rewrite: \texttt{"\{cfg['example']\}"} \\[6pt]

\texttt{[\{lang\} problem text---use its vocabulary]} \\
\texttt{\{q\_ref\}} \\

\bottomrule
\end{tabular}

\caption{System prompt used to produce the Mixed shell.}
\label{tab:mixed_prompt}
\end{table*}
\vspace{1em}

\subsection{Content-Preservation Judge Prompts}
\label{app:judge_prompts}

Tables~\ref{tab:judge_translation_prompt} and
\ref{tab:judge_mixed_prompt} reproduce the user-message rubrics used by
GPT-4o to evaluate the Translated and Mixed renderings, respectively. We use
no separate system message for these evaluations. The corresponding English
trajectory and rendered response are inserted into the prompt placeholders,
and \texttt{<LANG>} is replaced with the target language. The judge is
instructed not to solve the problem or assess the mathematical correctness of
the English trajectory, but to compare the two responses, score each
criterion independently, and return JSON only.

\begin{table*}[t]
\centering
\begin{tabular}{@{}p{1\textwidth}@{}}
\specialrule{1pt}{0pt}{0pt}
\rowcolor{gray!18}[0pt][0pt]
\textbf{  User Prompt} \\
\hline
\hline
\addlinespace[2pt]
You are auditing whether a \texttt{<LANG>} translation preserves the
reasoning content of an English math solution. Do NOT solve the problem or
evaluate whether the solution itself is mathematically correct. Treat the
English solution as the sole reference and compare ONLY the two texts.
\\[6pt]
\texttt{[EN solution]} \\
\texttt{<english solution>} \\[6pt]
\texttt{[<LANG> translation]} \\
\texttt{<translated solution>} \\[6pt]
Rate each dimension independently. Output valid JSON only. \\[6pt]
\textbf{1. steps\_preserved (1--5)} \\
Are all reasoning steps present in the same order, with no step added,
omitted, or reordered? \\[3pt]
\textbf{2. logic\_preserved (1--5)} \\
Are premise-to-conclusion relations, conditions, causal links, and logical
connectives faithfully preserved? \\[3pt]
\textbf{3. math\_preserved (1--5)} \\
Are all numbers, variables, operators, equations, inequalities, units, and
mathematical relations preserved exactly? \\[3pt]
\textbf{4. information\_delta} \\
Identify whether any non-mathematical information is added or omitted.
Return a list of ``added''/``omitted'' (empty if neither). \\[3pt]
\textbf{5. semantic\_equivalence (1--5)} \\
Would a reader who sees only the \texttt{<LANG>} text recover the same
reasoning, intermediate conclusions, and final conclusion as a reader of the
English text? \\
\quad (Each 1--5 dimension includes explicit 5-to-1 anchors.) \\[6pt]
Output exactly: \\
\texttt{\{"steps":n,"logic":n,"math":n,"info\_delta":[],"sem":n\}} \\
\bottomrule
\end{tabular}
\caption{User-message rubric used by GPT-4o to evaluate content preservation
in the Translated shell.}
\label{tab:judge_translation_prompt}
\end{table*}
\vspace{1em}

\begin{table*}[t]
\centering
\begin{tabular}{@{}p{1\textwidth}@{}}
\specialrule{1pt}{0pt}{0pt}
\rowcolor{gray!18}[0pt][0pt]
\textbf{  User Prompt} \\
\hline
\hline
\addlinespace[2pt]

You are auditing whether a LANGUAGE-MIXED math solution was produced
correctly. The mixing procedure takes an English solution and rewrites it so
that ONLY the named entities and key terms quoted from the problem are
rendered in \texttt{<LANG>}, while ALL reasoning, connectives, and
scaffolding stay in English. Numbers and equations must be unchanged. Do NOT
solve the problem or judge whether the math is correct---compare ONLY the two
texts against this procedure. \\[6pt]

\texttt{[EN solution]} \\
\texttt{<english solution>} \\[6pt]

\texttt{[Language-mixed (<LANG>) version]} \\
\texttt{<language-mixed solution>} \\[6pt]

Rate each dimension independently. Output valid JSON only. \\[6pt]

\textbf{1. entity\_localization (1--5)} \\
Are the problem's named entities and key content terms rendered in
\texttt{<LANG>} rather than left in English? \\[3pt]

\textbf{2. scaffolding\_english (1--5)} \\
Is the reasoning, connective, and explanatory text correctly kept in English
and not translated into \texttt{<LANG>}? \\[3pt]

\textbf{3. math\_preserved (1--5)} \\
Are all numbers, variables, operators, equations, inequalities, units, and
mathematical relations preserved exactly? \\[3pt]

\textbf{4. semantic\_equivalence (1--5)} \\
Would a reader of the mixed text recover the same reasoning, intermediate
conclusions, and final answer as a reader of the English text? \\[3pt]

\textbf{5. information\_delta} \\
Identify whether any non-mathematical information is added or omitted.
Return a list of ``added''/``omitted'' (empty if neither). \\

\quad (Each 1--5 dimension includes explicit 5-to-1 anchors.) \\[6pt]

Output exactly: \\
\texttt{\{"entity\_loc":n,"scaffold\_en":n,"math":n,%
"info\_delta":[],"sem":n\}} \\

\bottomrule
\end{tabular}

\caption{User-message rubric used by GPT-4o to evaluate content preservation
and mixing fidelity in the Mixed shell.}
\label{tab:judge_mixed_prompt}
\end{table*}
\vspace{1em}

\subsection{Content-Preservation Audit Results}
\label{app:audit_results}

Table~\ref{tab:content} reports the content-preservation audit results for
Translated and Mixed responses sampled from the constructed training
datasets, using their corresponding English trajectories as references.
The audit includes all pairs whose equation sets have a Jaccard similarity
below 0.9, together with a random sample of the remaining pairs. Results
are organized by language and shell. All preservation dimensions score
near ceiling, with the lowest value (4.71) occurring for entity
localization in the German Mixed rendering. The judge identifies no added
or omitted information in at least 98.8\% of evaluated pairs in every
condition.

\begin{table}[t]
\centering
\scriptsize
\setlength{\tabcolsep}{2.8pt}
\renewcommand{\arraystretch}{1.05}
\begin{tabular*}{\columnwidth}{@{\extracolsep{\fill}}l r cccc c@{}}
\toprule
\multicolumn{7}{@{}l}{\textit{Translated}} \\
Language & $N$ & Steps & Logic & Math & Sem. & No $\Delta$ \\
\midrule
Korean  & 1,305 & 4.98 & 4.98 & 4.99 & 4.98 &  99.7 \\
Chinese & 1,874 & 5.00 & 4.99 & 5.00 & 4.99 &  99.9 \\
German  &   857 & 4.99 & 5.00 & 5.00 & 4.99 & 100.0 \\
\textbf{All} & \textbf{4,036} & \textbf{4.99} & \textbf{4.99} & \textbf{5.00} & \textbf{4.99} & \textbf{99.9} \\
\midrule
\multicolumn{7}{@{}l}{\textit{Mixed}} \\
Language & $N$ & Entity & Scaffold & Math & Sem. & No $\Delta$ \\
\midrule
Korean  &   542 & 4.93 & 5.00 & 4.99 & 4.98 & 99.1 \\
Chinese &   695 & 4.93 & 5.00 & 5.00 & 4.99 & 99.6 \\
German  &   492 & 4.71 & 5.00 & 5.00 & 4.99 & 98.8 \\
\textbf{All} & \textbf{1,729} & \textbf{4.87} & \textbf{5.00} & \textbf{5.00} & \textbf{4.99} & \textbf{99.2} \\
\bottomrule
\end{tabular*}
\caption{
Content-preservation audit of the Translated and Mixed renderings.
Steps, Logic, Math, Sem., Entity, and Scaffold are rated on a 1--5 scale
by a GPT-4o judge. Steps and Logic assess preservation of reasoning steps
and logical relations; Math assesses mathematical equivalence; Sem. assesses
overall semantic equivalence; and Entity and Scaffold assess target-language
entity rendering and English scaffold retention, respectively. No $\Delta$
denotes the percentage of pairs for which the judge identifies no information
as added or omitted.
}
\label{tab:content}
\end{table}

\subsection{Human Evaluation of Shell Rendering Quality}
\label{app:human_evaluation}

To complement the GPT-4o content-preservation audit, we conduct a human evaluation of Korean and Chinese shell renderings. For each language, two native annotators independently evaluate 100 samples (50 Translated and 50 Mixed) on a 1--5 scale, focusing on subtle mistranslations and entity-rendering errors.

Annotators identify no errors in the sampled Translated responses, while three Mixed samples (3\% of Mixed samples) contain entity-rendering errors. However, their reasoning steps, equations, and final answers remain correct. Overall, 98\% of samples receive scores of at least 4 from both annotators, providing additional support for the quality of the LLM-rendered supervision.

\section{Renderer-Controlled Comparison}
\label{app:renderer_control}

Because the English shell retains the original Qwen3-32B response whereas
Translated and Mixed are rendered by Gemini-2.5-Flash, renderer identity is
confounded with shell realization in the main comparison. We therefore add
an \textit{English-Rerendered} (ER) control, in which Gemini rewrites each
verified trajectory in English under the same content-preserving constraints
(Table~\ref{tab:en2en_prompt}). ER is substantially rewritten at the prose
level, with corpus BLEU~\citep{papineni-etal-2002-bleu} of 67.5 overall and
41.8 on prose alone (equations, digits, and markup excluded). Comparing
English with ER tests the effect of rerendering while holding language
fixed, while comparing ER with Translated provides a renderer-matched test
of the English--Translated contrast.

We conduct this comparison for (Qwen, Korean), the setting with the largest
observed English--Translated gap in the main analysis. The ER condition uses the same
7,215 Korean training items as the three main shell conditions, with each
rerendered response verified to retain the gold final answer. Training and
evaluation follow the protocols described in
Sections~\ref{sec:models_training} and~\ref{sec:evaluation}, with two seeds
per condition. Table~\ref{tab:renderer_control} reports seed-averaged
accuracy.

We find no detectable renderer effect in this (Qwen, Korean) control:
English and ER are not separated at either difficulty level, with identical
Medium accuracy ($\Delta$acc $= 0.0$ pp, 95\% CI $[-6.0,+6.0]$). In
contrast, with Gemini fixed as the renderer, ER exceeds Translated on Medium
by 9.6 pp (95\% CI $[+3.2,+16.0]$), matching the original
English--Translated point estimate. Renderer identity therefore does not appear to account for the (Qwen, Korean) Medium-difficulty gap, while the Low-difficulty contrast remains
inconclusive.

\begin{table}[t]
\centering
\scriptsize
\renewcommand{\arraystretch}{1.05}
\begin{tabular}{@{}p{\columnwidth}@{}}
\specialrule{1pt}{0pt}{0pt}
\rowcolor{gray!18}[0pt][0pt]
\textbf{System Prompt} \\
\hline
\hline
\addlinespace[2pt]
You are a professional editor. \\
Rewrite the following math solution in natural English. \\[3pt]
\textbf{Rules:} \\
1. This is REWRITING, not solving---reproduce the same reasoning steps
in the same order; never re-solve, correct, or add steps. \\
2. Keep ALL numbers, calculations, LaTeX (\texttt{\$...\$},
\texttt{\$\$...\$\$}, \texttt{\textbackslash frac}), and markdown symbols
(\texttt{\#\#\#}, \texttt{**}, \texttt{---}) EXACTLY as they are; rewrite
only the prose around them. \\
3. Render all prose in fluent English, keeping person names as they
appear in the original. \\
4. End with exactly \texttt{'The answer is N.'} and output ONLY the
rewritten solution, no preamble. \\
\bottomrule
\end{tabular}
\caption{System prompt used to produce the English-Rerendered (ER)
control.}
\label{tab:en2en_prompt}
\end{table}
\begin{table}[t]
\centering
\small
\begin{tabular}{lcc}
\toprule
Shell & Low & Med. \\
\midrule
English (original)  & \textbf{88.8} & \underline{35.2} \\
ER (Gemini)         & \underline{87.2} & \underline{35.2} \\
Translated (Gemini) & 82.8 & 25.6 \\
Mixed (Gemini)      & 85.2 & \textbf{40.8} \\
\bottomrule
\end{tabular}
\caption{Seed-averaged final-answer accuracy (\%) on the Low and Medium
subsets of PolyMath for the renderer-controlled comparison (Qwen, Korean).
Parentheses indicate the source of each rendering: the original Qwen3-32B
teacher response or a Gemini-2.5-Flash rendering. The highest value in
each column is shown in \textbf{bold} and the second-highest is
\underline{underlined}.}
\label{tab:renderer_control}
\end{table}

\section{Training and Evaluation Details}
\label{app:training_details}

\subsection{Training Hyperparameters}
\label{app:training_hyperparameters}

We perform full-parameter fine-tuning using TRL's
\texttt{SFTTrainer} with DeepSpeed ZeRO-2. We use the 8-bit AdamW
optimizer with $\beta_1=0.9$, $\beta_2=0.999$, a learning rate of
$1\times10^{-5}$, cosine decay, and a warmup ratio of 0.05. Weight
decay is set to 0, and the maximum gradient norm is 1.0. All student
models and shell conditions use the same optimization settings. Each
student--language--shell condition is trained with two random seeds,
42 and 1234.

All experiments are conducted using two NVIDIA RTX 6000 Ada Generation
GPUs. The per-device batch size is 2 with 16 gradient-accumulation
steps, yielding an effective global batch size of 64. We set
$\texttt{max\_steps}=\lceil 3N/64\rceil$, where $N$ is the number of
retained training examples for each language. This results in 339
steps for Korean, 336 for Chinese, and 338 for German, corresponding
to approximately three passes over each language-specific dataset.
The maximum sequence length is 2,048 tokens. We disable sequence
packing and apply no truncation; all retained examples fit within this
limit. Training uses bfloat16 precision with gradient checkpointing.
Loss is computed only over response tokens, with prompt tokens masked,
and includes the terminal EOS token.

\subsection{Prompt Templates}
\label{app:prompt_template}

We use the same language-specific prompt prefix for fine-tuning,
compatibility measurement, and downstream evaluation. During fine-tuning,
the shell-specific response is appended after the final field label and only
the response tokens contribute to the training loss. For compatibility, we
compute the likelihood of the corresponding shell response under the same
prefix. At evaluation time, generation begins immediately after the final
field label. Table~\ref{tab:training_prompt} provides the complete templates,
where \texttt{\{question\}} is replaced with the target-language problem.

\begin{table}[t]
\centering
\footnotesize
\setlength{\tabcolsep}{2pt}
\renewcommand{\arraystretch}{1.05}

\begin{tabular}{
    @{}
    p{0.17\columnwidth}
    p{0.77\columnwidth}
    @{}
}
\toprule
\textbf{Language} & \textbf{Prompt Template} \\
\midrule
\addlinespace[2pt]

Korean &
\begin{minipage}[t]{\linewidth}
\begin{CJK}{UTF8}{mj}
다음 수학 문제를 풀이 과정과 함께 풀어라.

\smallskip
문제: \texttt{\{question\}}

\smallskip
풀이:
\end{CJK}
\end{minipage}
\\[4pt]

\midrule
\addlinespace[2pt]

Chinese &
\begin{minipage}[t]{\linewidth}
\begin{CJK*}{UTF8}{gbsn}
解答下面的数学题，并写出解题过程。

\smallskip
题目: \texttt{\{question\}}

\smallskip
解答:
\end{CJK*}
\end{minipage}
\\[4pt]

\midrule
\addlinespace[2pt]

German &
\begin{minipage}[t]{\linewidth}
Löse die folgende Mathematikaufgabe und gib den Lösungsweg an.

\smallskip
Aufgabe: \texttt{\{question\}}

\smallskip
Lösung:
\end{minipage}
\\[2pt]

\bottomrule
\end{tabular}

\caption{Language-specific prompt templates shared across fine-tuning,
compatibility measurement, and downstream evaluation.}
\label{tab:training_prompt}
\end{table}

\subsection{Evaluation Details}
\label{app:eval_details}

\paragraph{Answer extraction.}
We use the official PolyMath problems and gold answers, but score them with
our own rule-based checker instead of the official evaluation harness; no LLM
judge is involved. To extract the final answer from a generated response, we
try three patterns in order: (i) the last \verb|\boxed{}| expression;
(ii) the last line containing an answer marker in any of the four languages
(Korean, Chinese, German, or English), so a model is never penalized for
answering with a different language's marker; and (iii) if neither is found,
the last math expression or the last line of the response. If several markers
appear, we use the last one. Any answer that cannot be extracted or does not
match the gold answer is counted as incorrect. In practice, extraction never
fails: the no-prediction rate is 0.0\% for every shell condition at both
difficulty levels.

\paragraph{Answer matching.}
Extracted and gold answers are normalized by removing LaTeX formatting, then
compared by exact match, symbolic equivalence (SymPy), and numeric agreement
within $10^{-6}$. Equivalent forms such as \verb|1,000| vs.\ \verb|1000| and
\verb|\frac{1}{2}| vs.\ \verb|0.5| are treated as equal.

\paragraph{Comparability.}
We use a deterministic rule-based scorer rather than the official
LLM-judge-based evaluation, since a fixed, reproducible scoring rule is
better suited to controlled pairwise comparisons across shells. Absolute
accuracies are therefore not directly comparable to the PolyMath leaderboard,
but the same extraction and matching rules are applied to all shell
conditions, so the pairwise shell comparisons remain fair.

\paragraph{Language-proportion measurement.}
Korean and Chinese proportions are computed at the character level from
Unicode script ranges. Since German shares the Latin alphabet with English,
we instead count a word as German if it contains an umlaut or {\ss}, or
appears in a lexicon of 4{,}371 word types found only in the German training
responses (never in English, in at least three responses). This proxy is
noisier than the script-based measure, so we use it only to compare shells
within German. The final-answer marker line is retained in all languages and handled
identically across conditions; its contribution is therefore included
in the reported proportions.

\section{Robustness to Training-Run Variability}
\label{app:training_run_variability}

To further assess the robustness of our main results, we train two additional seeds per shell for Qwen in Korean and Chinese, focusing on the Mixed--Translated contrast on PolyMath Medium. We then compute 95\% confidence intervals using a two-level bootstrap that jointly resamples training seeds and evaluation items.

\begin{table}[t]
\centering
\small
\begin{tabular}{lcc}
\toprule
Language & $\Delta$Acc (pp) & 95\% CI \\
\midrule
Korean  & +9.8 & [+1.4, +18.2] \\
Chinese & +5.8 & [-1.4, +13.4] \\
\bottomrule
\end{tabular}
\caption{Four-seed Mixed--Translated accuracy differences on PolyMath Medium, with joint seed-and-item bootstrap 95\% confidence intervals.}
\label{tab:seed_robustness}
\end{table}

As shown in Table~\ref{tab:seed_robustness}, Mixed outperforms Translated across all four seeds in both languages. Compared with the original two-seed estimates, the four-seed mean differences decrease from 15.2 to 9.8 percentage points for Korean and from 9.2 to 5.8 for Chinese. The joint 95\% CI excludes zero for Korean but includes zero for Chinese. These results show a consistent direction of the Mixed--Translated effect across runs, while highlighting uncertainty in its magnitude.

\section{Cross-Benchmark Evaluation on MATH-500}
\label{app:math500}

To examine whether the observed shell effects are specific to PolyMath, we additionally evaluate the Qwen shell-trained models on MATH-500~\citep{lightman2024}, translated into Korean, Chinese, and German using Gemini-2.5-Flash with the question-translation prompt in Table~\ref{tab:question_translation_prompt}.

\begin{table}[t]
\centering
\small
\begin{tabular}{lccc}
\toprule
Shell & Ko & Zh & De \\
\midrule
English    & 68.3 & 72.8 & \textbf{71.6} \\
Translated & 64.6 & 69.4 & 69.6 \\
Mixed      & \textbf{69.4} & \textbf{73.4} & 69.0 \\
\bottomrule
\end{tabular}
\caption{Final-answer accuracy (\%) of Qwen shell-trained models on multilingual MATH-500, averaged over two seeds. The highest accuracy in each language is shown in bold.}
\label{tab:math500}
\end{table}

As shown in Table~\ref{tab:math500}, English and Mixed outperform Translated in Korean and Chinese, with nominal 95\% CIs excluding zero (English--Translated: +3.7/+3.4 pp; Mixed--Translated: +4.8/+4.0 pp, respectively). No English--Mixed contrast has a confidence interval excluding zero, nor does any shell contrast in German. These patterns are consistent with PolyMath, suggesting that the observed shell effects are not confined to a single evaluation benchmark.

\section{Additional Sequence-Length Analysis}
\label{app:sequence_length}

Table~\ref{tab:additional_length_analysis} extends the sequence-length
analysis to comparisons of Translated with both English and Mixed.
Panel A reports supervision-length ratios across all student--language
conditions. Panel B focuses on the five shell contrasts whose nominal
paired-bootstrap 95\% confidence intervals exclude zero in the main analysis.
For the four Medium contrasts, it reports both full-set and completed-only
results.

The supervision-length patterns do not align consistently with the observed
accuracy differences. In Chinese, Translated supervision traces are shorter
than Mixed ones under both student tokenizers (0.94 and 0.97), yet Mixed
outperforms Translated for Qwen on Medium problems, contrary to a simple
length-penalty account. Among the eight Medium regressions, seven slopes are
not distinguishable from zero at the nominal 95\% level. The sole nominal
exception is negative, opposite to the positive association expected if
greater relative Translated length systematically increased its accuracy
disadvantage. Restricting the analysis to items completed within the token
budget increases the accuracy gap in all four Medium contrasts, indicating
that generation-budget truncation does not account for these differences.
Together, these results provide no consistent evidence that relative sequence
length explains the observed Translated-shell disadvantages.

\begin{table}[t]
\centering
\footnotesize
\renewcommand{\arraystretch}{1.08}

\begin{tabular*}{\columnwidth}{
    @{\extracolsep{\fill}}
    llrr
    @{}
}
\toprule
\multicolumn{4}{@{}l@{}}{
    \textbf{A. Supervision-length ratios}
} \\
Student & Lang. & Tr./En. & Tr./Mx. \\
\midrule
\multirow{3}{*}{Qwen}
& Ko & 1.31 & 1.18 \\
& Zh & 1.06 & 0.94 \\
& De & 1.23 & 1.16 \\
\midrule
\multirow{3}{*}{Gemma}
& Ko & 1.15 & 1.08 \\
& Zh & 1.04 & 0.97 \\
& De & 1.11 & 1.08 \\
\bottomrule
\end{tabular*}

\vspace{5pt}

\begin{tabular*}{\columnwidth}{
    @{\extracolsep{\fill}}
    lllrr
    @{}
}
\toprule
\multicolumn{5}{@{}l@{}}{
    \textbf{B. Generated-length regressions for detected effects}
} \\
Lang./Diff. & Contrast & Subset
& $\Delta$Acc
& $\beta$ ($p$) \\
\midrule
Ko Low
& En--Tr
& All
& $+6.0$
& $+0.097$ (.26) \\
\midrule
\multirow{4}{*}{Ko Med.}
& \multirow{2}{*}{En--Tr}
& All
& $+9.6$
& $+0.058$ (.16) \\
&
& Comp.
& $+13.3$
& $+0.096$ (.35) \\
\cmidrule(lr){2-5}
&
\multirow{2}{*}{Mx--Tr}
& All
& $+15.2$
& $+0.075$ (.13) \\
&
& Comp.
& $+16.3$
& $+0.044$ (.68) \\
\midrule
\multirow{4}{*}{Zh Med.}
& \multirow{2}{*}{En--Tr}
& All
& $+9.2$
& $-0.027$ (.43) \\
&
& Comp.
& $+9.5$
& $-0.349$ (.042) \\
\cmidrule(lr){2-5}
&
\multirow{2}{*}{Mx--Tr}
& All
& $+9.2$
& $+0.015$ (.74) \\
&
& Comp.
& $+12.5$
& $+0.066$ (.65) \\
\bottomrule
\end{tabular*}

\caption{Additional sequence-length analyses comparing Translated with
English and Mixed. En, Tr, and Mx denote the English, Translated, and Mixed
shells, respectively. Panel A reports supervision-length ratios under each
student tokenizer, where Tr./En.\ and Tr./Mx.\ are the ratios of Translated
to English and Mixed supervision lengths. Panel B reports regressions for
the five Qwen contrasts whose nominal 95\% confidence intervals exclude zero.
For comparison shell $C$, $\Delta$Acc $= C -$ Translated, and $\beta$ uses
the Translated-to-$C$ generated-length ratio. Comp.\ includes only items for
which both models complete generation within the token budget.}
\label{tab:additional_length_analysis}
\end{table}

\section{AI Assistant Usage}

We have used Claude Code during the development of our research work.

\end{document}